\documentclass[journal]{IEEEtran}

\usepackage{cite}

\ifCLASSINFOpdf
  \usepackage[pdftex]{graphicx}
  \usepackage{graphics}
  \DeclareGraphicsExtensions{.pdf,.jpeg,.png}
\else
  \usepackage[dvips]{graphicx}
  \usepackage{graphics}
  \DeclareGraphicsExtensions{.eps}
\fi

\usepackage{amsmath}
\usepackage{amssymb,amsthm}
\usepackage{algorithm}
\usepackage{algorithmic}

\usepackage{threeparttable}
\usepackage{array}
\usepackage{multirow}
\usepackage{colortbl,booktabs}

\ifCLASSOPTIONcompsoc
  \usepackage[caption=false,font=normalsize,labelfont=sf,textfont=sf]{subfig}
\else
  \usepackage[caption=false,font=footnotesize]{subfig}
\fi

\ifCLASSOPTIONcaptionsoff
  \usepackage[nomarkers]{endfloat}
  \let\MYoriglatexcaption\caption
  \renewcommand{\caption}[2][\relax]{\MYoriglatexcaption[#2]{#2}}
\fi

\usepackage{url}

\usepackage{pgfplots}
\pgfplotsset{compat=1.13}

\usepackage{afterpage}
\usepackage{lscape}

\begin{document}

\title{Apparel-invariant Feature Learning for Person Re-identification}

\author{Zhengxu~Yu, Yilun~Zhao, Bin~Hong,
        Zhongming~Jin, 
        Jianqiang~Huang, \\Deng~Cai,~\IEEEmembership{Member,~IEEE}, Xiaofei He,~\IEEEmembership{Senior Member,~IEEE} and Xian-Sheng Hua,~\IEEEmembership{Fellow,~IEEE}
\thanks{Z.~Yu, D.~Cai and X.~He are with the State Key Laboratory of CAD\&CG, College of Computer Science, Zhejiang University, Hangzhou, Zhejiang 310058, China (emails: yuzxfred@gmail.com; dengcai@gmail.com; xiaofeihe@zju.edu.cn).}%
\thanks{Y.~Zhao is with Zhejiang University, Hangzhou, Zhejiang 310058, China (emails: zhaoyilun@zju.edu.cn).}%
\thanks{Z.~Jin, B.~Hong, J.~Huang and X.-S.~Hua are with Alibaba Group, Hangzhou, Zhejiang, China (emails: zhongming.jinzm@alibaba-inc.com; bin\_hong@zju.edu.cn; jianqiang@alibaba-inc.com; xiansheng.hxs@alibaba-inc.com).}}

\ifCLASSOPTIONpeerreview
\markboth{IEEE Transactions on Image Processing}{Yu \MakeLowercase{\textit{et al.}}: Apparel-invariant Feature Learning for Apparel-changed Person Re-identification}
\fi

\maketitle

\begin{abstract}
With the rise of deep learning methods, person Re-Identification (ReID) performance has been improved tremendously in many public datasets. However, most public ReID datasets are collected in a short time window in which persons' appearance rarely changes. In real-world applications such as in a shopping mall, the same person's clothing may change, and different persons may wearing similar clothes. All these cases can result in an inconsistent ReID performance, revealing a critical problem that current ReID models heavily rely on person's apparels. Therefore, it is critical to learn an apparel-invariant person representation under cases like cloth changing or several persons wearing similar clothes. In this work, we tackle this problem from the viewpoint of invariant feature representation learning. The main contributions of this work are as follows. (1) We propose the semi-supervised Apparel-invariant Feature Learning (AIFL) framework to learn an apparel-invariant pedestrian representation using images of the same person wearing different clothes. (2) To obtain images of the same person wearing different clothes, we propose an unsupervised apparel-simulation GAN (AS-GAN) to synthesize cloth changing images according to the target cloth embedding. It's worth noting that the images used in ReID tasks were cropped from real-world low-quality CCTV videos, making it more challenging to synthesize cloth changing images. We conduct extensive experiments on several datasets comparing with several baselines. Experimental results demonstrate that our proposal can improve the ReID performance of the baseline models.
\end{abstract}

%
\IEEEpeerreviewmaketitle
\section{Introduction}
Given an image of the person-of-interest under one camera, person re-identification (ReID) aims to annotate the target person by matching it with historical images captured by other cameras in the same multi-camera network. ReID has attracted a lot of attention from the community in recent years due to its broad application prospects in new retail business and public security, such as crime prevention \cite{wang2013intelligent}, finding elder and children \cite{zhao2017spindle} and person activity analysis \cite{loy2009multi}.
\begin{figure}[!htp]
\centering
\includegraphics[width=0.50\textwidth]{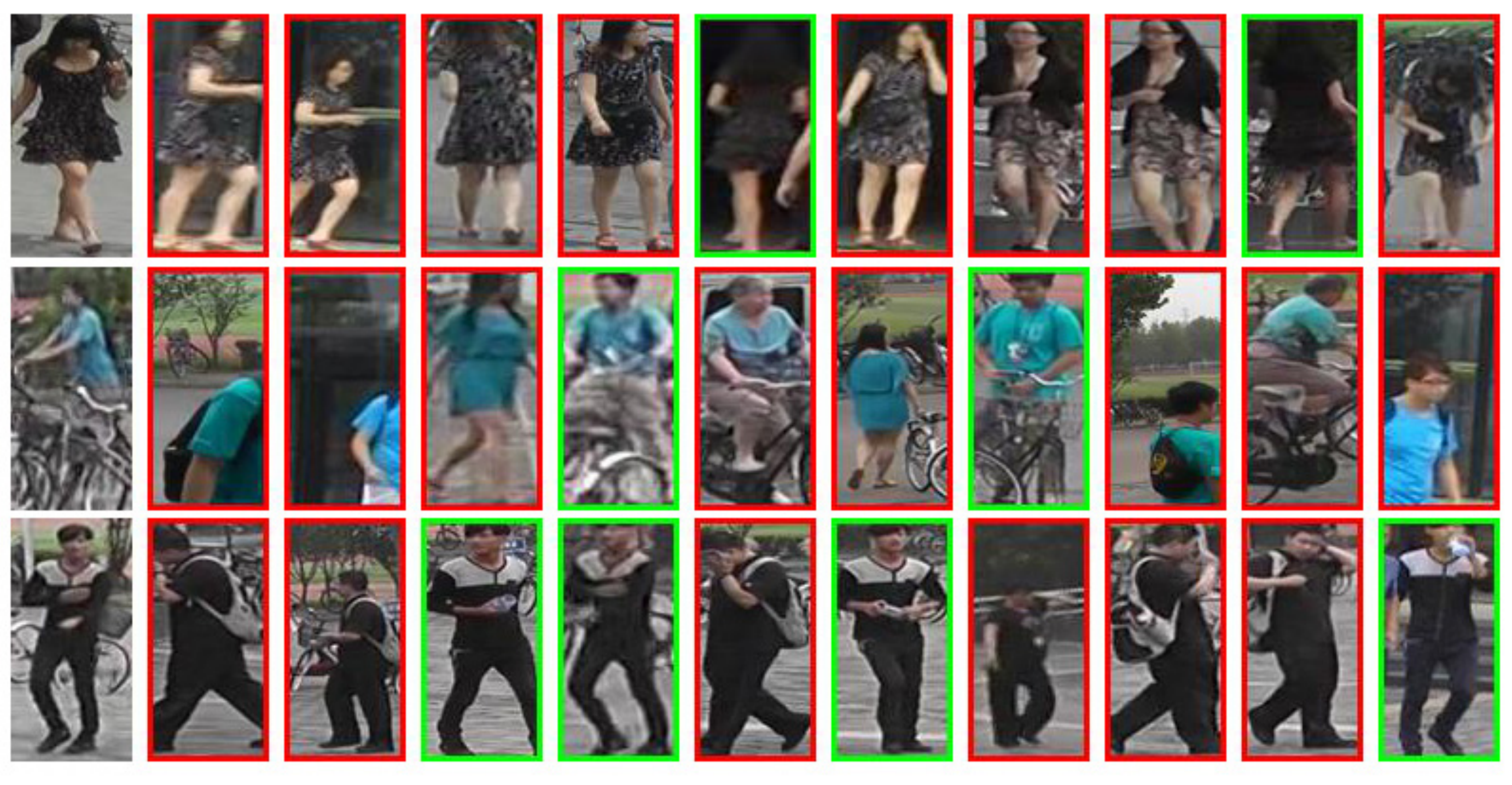}
\caption {\textbf{Bad case analysis of previous work.} Each column is a top-10 ReID list obtained using the ReID method proposed by Hermans et al. \cite{hermans2017defense}. The first left person image of each column is the target person sampled from PAVIS dataset \cite{Barbosa:reid12}. The red box indicates an incorrect match, and the green one indicates a correct match. It is clear that the ReID model heavily relied on the cloth information.}
\label{fig:sample-market}
\end{figure}

Person ReID is challenging because the images ReID used are cropped from low quality CCTV videos, which contain intensive variations like illumination, human pose, camera viewpoint, occlusion, and clothing change. Hence, learning a discriminative and robust person representation under such variations is one of the most fundamental problems in person ReID tasks. Early person ReID methods usually use hand-crafted discriminative features constructed with low-level features such as color and texture information of the images \cite{liao2015person,chen2016similarity,zhao2013person}. However, in cases of illumination change or cases like several persons wearing similar clothes, these low-level features become unreliable \cite{wu2017robust}. More recently, many deep-learning based methods were proposed \cite{li2014deepreid,ahmed2015improved,zheng2017person,yu2017cross,almazan2018re,li2017person,zhao2017deeply,zheng2017pose,xiao2017joint}, which applied deep convolutional networks to learn identity-sensitive and view-insensitive embeddings based on image classification and metric learning. Meanwhile, there are also many unsupervised methods were proposed to make use of all possible training data \cite{fan2017unsupervised,lv2018unsupervised,2018arXiv180309786W,DBLP:journals/corr/abs-1711-07027,yu2017cross}.

Although these recently proposed ReID methods greatly improved the ReID performance in several public datasets, there is a critical problem that most of these ReID models are easily misled by the cloth feature which they rely on. This problem leeds to an unstable performance in real-world long-term ReID applications. We provided a case study in \textbf{Fig.\ref{fig:sample-market}} to help understand the most discriminative features the previous ReID methods relied on is the target person's apparel features, especially cloth features. In most of the existing datasets, images are taking in a short time window like in several days, which people rarely change their clothing. Thus, the same person's appearance is roughly consistent in different images. Unlike that, the real-world ReID systems (e.g., in a shopping mall) are usually continuously online after deployment. In this case, there may be a long interval between the two shots of the target person, during which there will be many environmental variations, as well as cloth changing and different persons wearing similar clothes. Hence, in real-world applications, the previous ReID models which heavily relied on the apparel features would have substantial performance degradation. 

In this paper, we consider cases where the most consistent feature (apparel) is no longer reliable due to apparel changing or different people wearing similar clothes cases. These apparel-changed person ReID problem widely existed in long term person ReID applications like criminal investigation and person activity analysis. Hence, it is crucial to find an apparel-invariant feature learning method to learn a more discriminative feature in this apparel-changed scenario. 

To solve this apparel-changed ReID problem, we propose the semi-supervised Apparel-invariant Feature Learning (AIFL) framework (\textbf{Sec. \ref{sec:pre-train}}) to learn an apparel irrelevant feature representation. To the best of our knowledge, there is no existing dataset contains cloth IDs for each person. Therefore, we first propose a GAN based apparel-simulation image generator AS-GAN(\textbf{Sec. \ref{sec:cloth-changer}}) to generated synthetic apparel changed images for persons in the training dataset. The simplest way to use the AS-GAN is to create a new training dataset by mixing the images generated by AS-GAN and the original images. It could be considered as a data augmentation method. However, the intruding of synthetic images changed the data distribution, which might deviate from the original distribution. As shown in the experiment results in \textbf{Table. \ref{tab:msmt17}}, directly use the synthetic images is not an effective way.

We categorize the contribution of this work into threefold:
\begin{enumerate}
\item We propose a novel semi-supervised Apparel-invariant Feature Learning (AIFL) framework in this work to learn a discriminative feature embedding in cases like apparel changing and the different persons wearing similar cloth cases, and thus improve the ReID performance.
\item To solve the problem of insufficient training data. We construct an apparel-simulation GAN (AS-GAN) to generate cloth changing images.
\item We conduct extensive experiments on datasets with significant cloth changing and the different person wearing similar cloth cases, \textit{i.e.} PAVIS and MSMT17. Experimental results have shown that our proposal could effectively improve the ReID performance of baseline models.
\end{enumerate}

\section{Related Works}
\subsection{Person ReID under intensive appearance variations}

The negative effects of appearance variation on deep-learning based ReID models have been recognized recently. This appearance variation not only includes the external changes like illuminations and occlusion but also included the person's appearance changes like apparel and pose. A number of models have proposed to address these problems\cite{su2017pose,DBLP:journals/corr/abs-1711-07027,2017arXiv171202225Q,DBLP:journals/corr/abs-1711-08565}. As for the external variations, Deng \textit{et al.} \cite{DBLP:journals/corr/abs-1711-07027} proposed the similarity preserving generative adversarial network to mitigate the domain difference by conducting image style transferring. Wei \textit{et al.} \cite{DBLP:journals/corr/abs-1711-08565} proposed a person transfer model which transfer peoples from source domain into the target domain to mitigate the environment difference. As for the person's appearance changes, many pose-guided deep-learning ReID models\cite{su2017pose,2017arXiv171202225Q} were proposed recently. For instance, Qian \textit{et al.} \cite{2017arXiv171202225Q} proposed a pose-normalized image generator to generate different pose images for the same person. Peng \textit{et al.}\cite{lv2018unsupervised} propose an asymmetric multi-task dictionary learning method to transfer the view-invariant representation learned from source data to target data.

However, the person's apparel variations like cloth changing or different persons wearing similar clothes are rarely mentioned. The reason is easy to understand that the cloth is the most salient characteristic in most public ReID datasets. Thus most of RGB based ReID models heavily rely on the apparel similarity \cite{wu2017robust}. To solve this problem, Wu \textit{et al.}\cite{wu2017robust} proposed an RGB-D based person ReID model, which combined the estimated depth features with RGB appearance features to reduce the visual ambiguities of appearance caused by similar clothes. However, the RGB-D cameras they used are much more expensive than the regular RGB camera, which hindered its application in the real world. Distinct from them, our proposal is still an RGB-based model that reduces the visual ambiguities by learning a self-restraint apparel-invariant feature. Nevertheless, Wu \text{et al.}\cite{8237837} proposed a deep zero-padding model to minimize the modality gap between the RGB image and the IR image. The IR image they used does not contain color information. Our work is distinct from theirs in that we only use the RGB image without any extra information in the whole process.

\subsection{Image Generation}
One of the bottlenecks of deep-learning based person ReID models is the lack of training data. Hence, generating realistic images using GAN has received much interest recently. 

Zhu \textit{et al.}\cite{DBLP:journals/corr/abs-1710-07346} proposed a approach to generated realistic person images with different clothes. In their work, they used high-resolution images that are captured in a front view, and the size of the person in this image is nearly the same. They applied the openpose\cite{cao2017realtime,wei2016cpm} to generate human keypoint map to guild the cloth generation. Their work is related to our image generator. However, our image generator still distinct from their work in that we used the real-world low-resolution image, which contains intensive pose and viewpoint variations. 

Ma \textit{et al.}\cite{ma2018disentangled} proposed a two-stage reconstruction pipeline that learns a disentangled representation of the aforementioned image factors and generates new person images at the same time. The generated image of their work could have different apparel and pose, but the generated image does not guarantee the person ID consistency. Different from their work, the AS-GAN only changes the cloth part of the image to make sure the generated person image could be identified as the same person in the input image.

Qian \textit{et al.}\cite{2017arXiv171202225Q} proposed a pose-normalized image generator to generate the same person's images with a different pose. Inspired by their work, we propose our apparel-simulation GAN based upon conditional
GANs (cGANs)\cite{goodfellow2014generative}. However, our proposal is distinct from Qian's work because the condition information used in our image generator is different, and the network architecture is also different. We used cloth features extracted by a CNN model as the conditional code. This cloth feature extractor model is also trained in an unsupervised way. Meanwhile, the network structure is different. They used the 'U-net' structure to preserve the low-level information. However, in apparel-changed image generation, the original cloth's low-level information must be removed from the feature maps.
Pathak \textit{et al.}\cite{pathak2016context} proposes an unsupervised visual feature learning algorithm driven by context-based pixel prediction. They use the context information to generate the missing part of the input image. Inspired by their work, we erase the cloth part of the input image to eliminate the cloth's impact on the generated image. Our proposal is distinct from their work because the goal of their work is to repair the input image, but what we want is to change the content of the missing part.

\subsection{Unsupervised Learning}
Many unsupervised\cite{fan2017unsupervised, zhao2017person,ye2017dynamic,peng2016unsupervised,donahue2016adversarial} person ReID methods have been proposed recently. RGB-based Hand-crafted features\cite{gray2008viewpoint,matsukawa2016hierarchical,liao2015person} can be directly employed for unsupervised ReID. All these methods focused on feature design but ignored the information from the distribution of samples in the dataset. Moreover, the most commonly used feature in the hand-crafted feature based ReID methods is colour\cite{reid-survey}, especially the color of clothes. Consequently, these features will become unreliable in the presence of intensive apparel variations. Classical unsupervised learning methods such as auto-encoders\cite{bengio2009learning,bourlard1988auto} aim to learn useful feature representations by simply reconstructing it after a bottleneck. More related to our work, Donahue \textit{et al.}\cite{donahue2016adversarial} proposed a GAN based feature learning framework BiGAN, which learning feature representation by reconstructing the input image. However, our proposed AIFL framework is an auto-encoder based network which input and target image are two different images.

\section{Proposed Method}
\subsection{Problem Formulation and Overview}
Before presenting our method, we first introduce some basic notions and terminologies. 
Suppose a gallery set $\mathcal{G}$ contains $\mathit{N}$ cropped person images $\{ x_{k}\}_{k=1}^{N}$. They belong to $\mathit{N}$ different identities $1, 2, ..., \mathit{N}$. We first denote a feature extraction model as $\phi(\cdot;\theta)$, which is parameterized by $\theta$.
Given a query person $q\in\mathcal{Q}$, the identity of $q$, denote as $\mathit{id}(q)$, is determined as:

\begin{equation}
  \label{equ:problem-formulation-sim}
  k^{*} = \mathop{\arg\min_{k\in N}} d(\phi(q; \theta), \phi(x_{k}; \theta))
\end{equation}

Where $d(\cdot,\cdot)$ is some kind of distance function.

In the apparel-changed person ReID scenario, we step further to add a constraint to restraint the queried person should acquire the same ID when he/she is wearing different apparels:
\begin{equation}
\label{equ:qj=qj'}
    \mathit{id}(q_{j})=\mathit{id}(q_{j^{'}})
\end{equation}
Where $j, j^{'}$ is the two different apparels which person $q$ has worn.

By observing that:
\begin{equation}
\begin{aligned}
| d(\phi(q_{j}; \theta), \phi(x_{k}; \theta)) - d(\phi(q_{j^{'}}; \theta),\phi(x_{k}; \theta)) | \\
< d(\phi(q_{j}; \theta), \phi(q_{j^{'}}; \theta)) \\
\end{aligned}
\end{equation}
holds for $\forall k \in \mathit{N}$. 

To ensure \textbf{Eq. \ref{equ:qj=qj'}}, we need to make $d(\phi(q_{j}; \theta), \phi(q_{j^{'}}; \theta))$ as small as possible for $\forall q \in \mathcal{Q}$. Hence, the person ReID problem could be formulated as optimize the fellowing equation:
\begin{equation}
\label{equ:optimize-pro}
    \min \frac{1}{|\mathcal{Q}|}\sum_{q\in\mathcal{Q}}d(\phi(q_{j}; \theta), \phi(q_{j^{'}}; \theta))
\end{equation}

\subsection{Framework Overview}
\begin{figure}[!tp]
\includegraphics[width=0.50\textwidth]{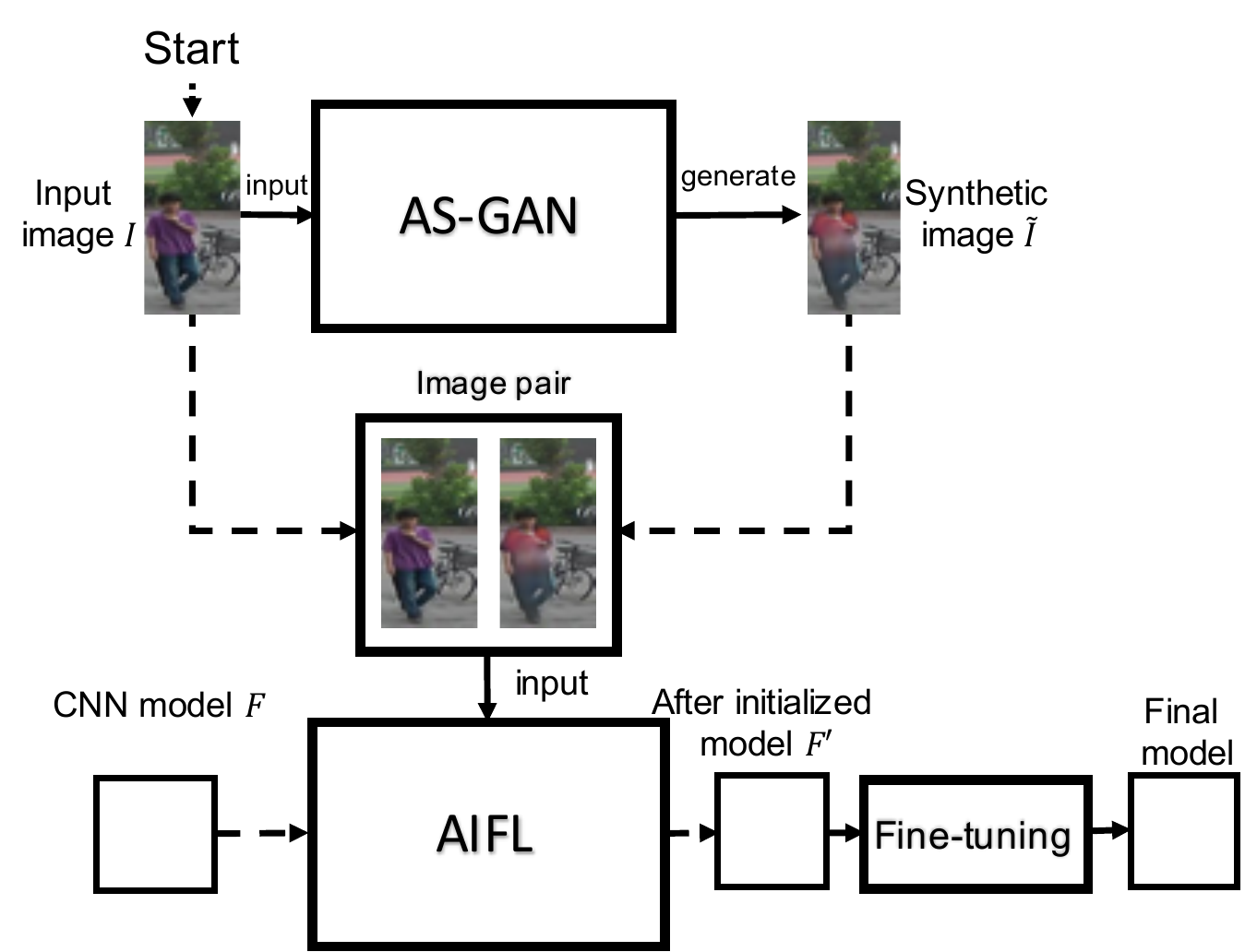}
\caption{\textbf{Overview} of our proposal. Our proposal is consists of two parts: 1) the apparel-simulation GAN(AS-GAN), 2) the apparel-invariant feature learning(AIFL) framework. The AS-GAN is design to generate synthetic cloth images. The AIFL framework is an auto-encoder based newtwork which is demonstrated in \textbf{Sec. \ref{sec:pre-train}}.}
\label{fig:2}
\end{figure}

As we mentioned above, there is no existing ReID dataset contains cloth label. Hence, we proposed a novel apparel-simulation image generator (AS-GAN) to generate the apparel-changed images using unlabelled irrelevant person images.

To make use of these unlabelled images, we introduce the core proposal of this paper -- the apparel-invariant feature learning (AIFL) framework.

The overview of our proposal have shown in \textbf{Fig.\ref{fig:2}}, our proposal has two key components, i.e., the apparel-invariant feature learning framework(AIFL) (Sec. \ref{sec:pre-train}), and the GAN based person image generation model(AS-GAN) (Sec. \ref{sec:cloth-changer}). 

During training, the original image and the generated image will be paired up and send into the AIFL framework. Give a CNN based feature extractor model $F$, we use the image pairs generated by AS-GAN to learn an apparel-invariant feature in AIFL. For each image pair (or an iteration), $F$ will be trained twice. We first set the original image and the synthetic image as target and input to train the $F$ for the first time. After that, we reverse the input and target and train the networks once more. By doing so, the apparel feature will be ignored by $F$ to yield a lower loss. We then add a CNN based classifier on the output of $F$ to conduct ReID application. To build up the connection between feature embedding and person IDs, we also conduct model fine-tuning on the ReID model $F$ and the classifier.

\subsection{Apparel-simulation Image Generator}
\label{sec:cloth-changer}
\begin{figure}[!tp]
\includegraphics[height=2in,width=0.5\textwidth]{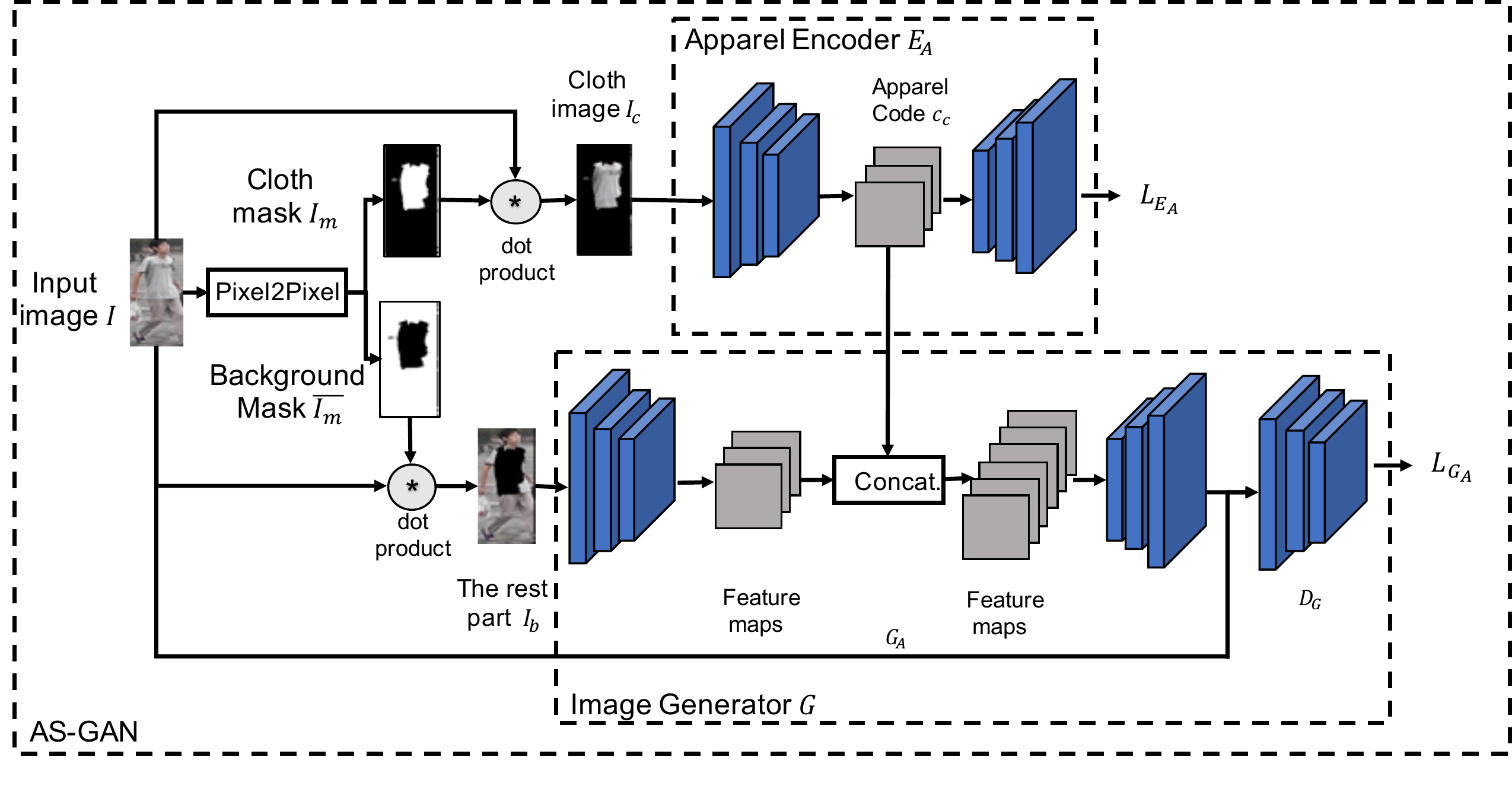}
\caption{\textbf{Schematic} of the apparel-simulation GAN. We first applied a pixel2pixel model to parse out the cloth mask, and then, extract the cloth code using an auto-encoder network. Then this cloth code will be used in the AS-GAN. During training, the cloth code is extracted from the input image. During generation, the cloth code is extracted from another randomly selected image in the same dataset. This figure only demonstrates the training process here.}
\label{fig:3}
\end{figure}
Inspired by \cite{2017arXiv171202225Q} and \cite{pathak2016context}, we constructed the apparel-simulation GAN (AS-GAN) to synthesize the same person's images conditional on apparel. This generator is formed by a GAN based image generator $\mathit{G}$ and an auto-encoder based apparel encoder $\mathit{E_{A}}$, as shown in \textbf{Fig.\ref{fig:3}}. Since the only thing we want to change is the cloth, it is natural to parse out the cloth and keep anything else unchanged. To know where the clothes are in the input image, we first trained a pixel2pixel \cite{pixel2pixel} model on HumanParsing-Dataset \cite{ATR} to generate cloth mask ($I_{m}$) for the input person image $I$. We then multiply the input image $I$ with the cloth mask $I_{m}$ to separate the input image into the cloth part ($I_{c}$) and the rest part ($I_{b}$). All the images used in this work are resized into 256x128 (height x width) in advance.

\textbf{Apparel encoder $\mathit{E_{A}}$.} We first constructed the auto-encoder based cloth encoder $\mathit{E_{A}}$ to generate a unique code for each cloth. The $\mathit{E_{A}}$ is constructed by a five-layer convolution neural network as encoder, following by a five-layer deconvolution neural network as the decoder. Each layer of the encoder is consists of a convolutional block (Conv2d block) contains a 2-dimensional convolutional (Conv2d) layer, a LeakyReLU layer and a BatchNorm layer. The negative slope is set to 0.2 in the LeakyReLU layer. Each layer of the decoder contains a 2-dimensional transpose convolutional (TransConv2d) block included a TransConv2d layer and the following ReLU layer and a BatchNorm layer. The encoder down-sampling the cloth image $I_c$ into a 512x8x4 (channel x height x width) feature maps (cloth code $c_{c}$). During training, we set the input cloth image $I_{c}$ as input and target at the same time. And we applied L1 loss in this auto-encoder network. The loss function $\mathcal{L}_{\mathit{E_{A}}}$ has shown below:
\begin{equation}
\label{equ:encoder}
\begin{split}
&\mathcal{L}_{\mathit{E_{A}}} = \parallel I_{c}-\mathit{E_{A}}(I_{c}) \parallel_{1}
\end{split} 
\end{equation}

Where $\parallel \cdot \parallel$ is the L1 loss, $I_{c}$ is a cloth image.

\begin{figure*}[!tp]
     \centering
     \includegraphics[ width=\textwidth]{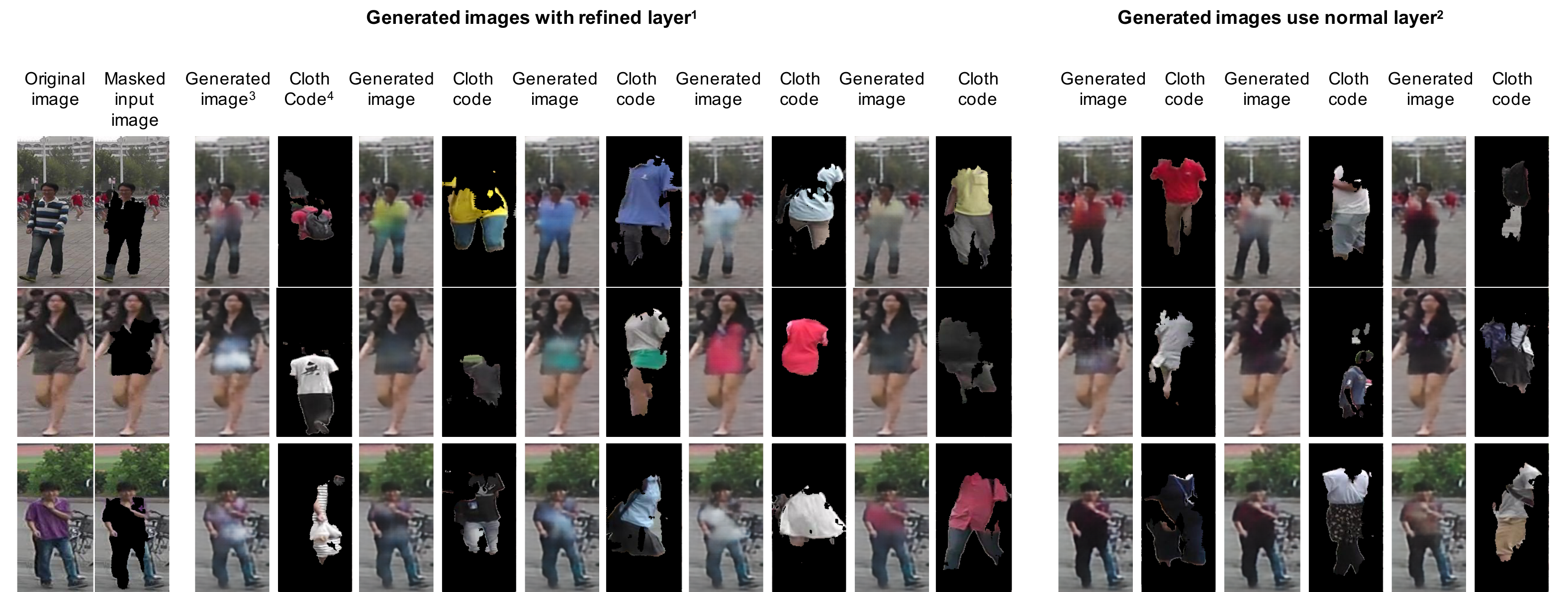}
\caption{\textbf{Samples generated by using AS-GAN}. We sampled 100 thousand images from the MARS dataset \cite{zheng2016mars}, and then generated five different clothes for each image using different cloth code sampled from the same dataset. As we can see, the most conspicuous part of the original cloth has changed according to the cloth code. Worth to mention, the generated cloth number is not optimal, more clothes could be generated using different cloth code. 1. Image generated by image generator with the second Conv2d block replaced by the refined layer, as shown in \textbf{Fig.\ref{fig:refinedlayer}}. 2. Image generated by image generator constructed by the normal 4x4 Conv2d block. 3. The AS-GAN generates the generated image with the randomly selected cloth code on the right. 4. The cloth code is generated by the apparel encoder $E_{A}$ with a randomly selected cloth as input.}
\label{fig:4}
\end{figure*}

\textbf{Image generator $\mathit{G}$.} To generate a more realistic image, we constructed the image generator $\mathit{G}$ based on conditional GANs. The $\mathit{G}$ has two components, a Generator $\mathit{G_{A}}$ and a Discriminator $\mathit{D_{G}}$, as shown in \textbf{Fig.\ref{fig:3}}. Given an input image $I_{q}$ of person $q$ and a random select cloth $I_c$. We can use $I_c$ as input to get the cloth code $c_{c}$ using the $\mathit{E_{A}}$. our image generator aims to synthesise a new image $\widetilde{I_{q}}$ under condition cloth code $c_{c}$, and ensure the identity consistency of $I_{q}$ and $\widetilde{I_{q}}$. The generator $\mathit{G_{A}}$ consists of 5 down-sampling block, followed by five up-sampling blocks. Except for the second down-sampling block, all the other down-sampling blocks contain a 4x4 kernel size Conv2d layer with stride 2 and padding 1 followed by a LeakyReLU layer and a BatchNorm layer. The negative slope of these LeakyReLU layers is set to 0.2. And each up-sampling block includes a TransConv2d layer followed by a ReLU layer and a BatchNorm layer. The apparel code $c_{c}$ generated by $\mathit{E_{A}}$ is concatenated with the output of the last down-sampling block of $\mathit{G_{A}}$. We then send the mixture into a 1x1 Conv2d layer and the following up-sampling blocks.

To eliminate the information of the original apparel in $I_{q}$, inspired by \cite{pathak2016context}, we use the complement $\bar{I_{m}}$ of the cloth mask $I_{m}$ to set the cloth part of the input image $I$ as 0. We denote this image without cloth part as $I_{b}$.

\begin{figure}
    \centering
    \includegraphics[width=0.40\textwidth]{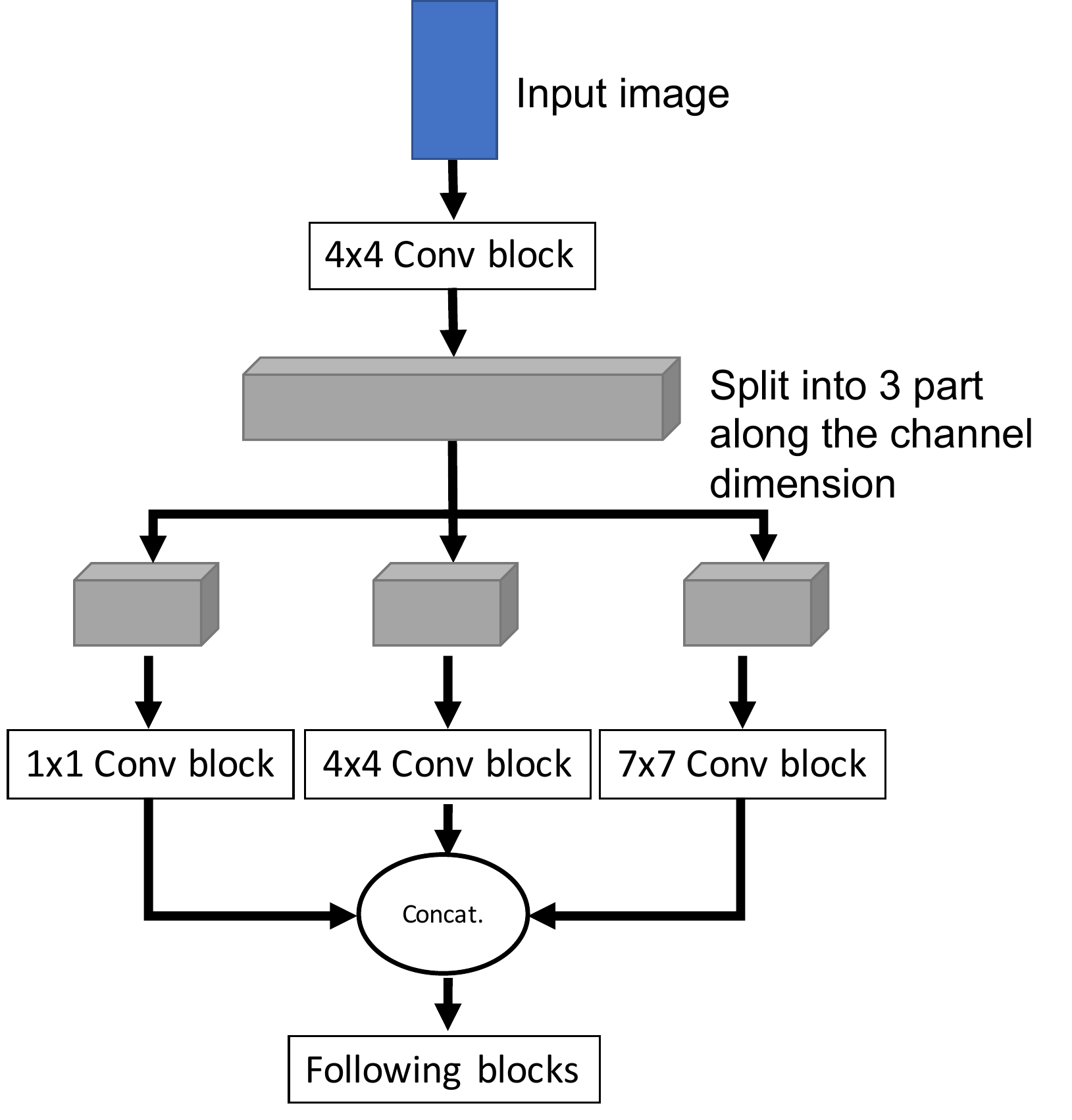}
    \caption{\textbf{The refined layer}. The output of the first layer is split into 3 part, and each part feed into a sub-block. The sub-blocks is consist of three Conv2d blocks with kernel size 1x1, 4x4 and 7x7 respectively. The output of these sub-block will be concatenated and feeded into the following layers.}
    \label{fig:refinedlayer}
\end{figure}

The second down-sampling block consists of three different kernel size Conv2d layers (names the refined layer). Each one followed by a LeakyReLU layer and a BatchNorm layer, as shown in \textbf{Fig.\ref{fig:refinedlayer}}. The output of the first layer is divided along the channel into three equal parts. Each part feed into the different kernel size Conv2d block. The generated images compare with only use one 4x4 Conv2d block as the second down-sampling block has shown in \textbf{Fig.\ref{fig:4}}.

The discriminator $\mathit{D_{G}}$ consists of 6 Conv2d blocks and a linear layer. Each Conv2d block contains a Conv2d layer and a BatchNorm layer and a LeakyReLU layer. This discriminator aims to discriminate between real data samples from the generated samples and thus helps to improve the quality of produced images.

\textbf{Work Flow.} As shown in \textbf{Fig.\ref{fig:3}}.
During \textit{training}, we first use the cloth mask $I_{m}$ generated by the pixel2pixel model(as discussed above) to separate the input image $I$ into cloth image $I_{c}$ and the rest part image $I_{b}$. We then extract the cloth code $c_{c}$ using cloth image $I_{c}$. By setting the input image as the target image at the same time, this image generator could learn how to use the cloth code to reconstruct the original image. To help $\mathit{G_{A}}$ reconstruct the cloth in the right place, $I_{c}$ was randomly flipped and cropped before feeding into apparel encoder $\mathit{E_{A}}$. Because of the cloth code only contains the cloth information and the only thing missing in the rest part image is the cloth, the image generator will learn to use the cloth code to reconstruct the cloth part while using the rest part image to reconstruct other things like human body and background.
The objective of the AS-GAN could be expressed as:
\begin{equation}
\label{equ:as}
\begin{split}
&\mathcal{L}_{\mathit{D_{G}}} = \min_{\mathit{G_{A}}} \max_{\mathit{D_{G}}} \mathbb{E}_{I, c_{c}}\{\log \mathit{D_{G}}(I) + \log(1-\mathit{D_{G}}(\mathit{G_{A}}(I_{b},c_{c}))\} \\
&\mathcal{L}_{\mathit{G_{A}}} = \parallel I - \mathit{G_{A}}(I_{b},c_{c}) \parallel_{1} + \lambda_{\mathit{D_{G}}} \mathcal{L}_{\mathit{D_{G}}} 
\end{split}
\end{equation}
Where the $\mathcal{L}_{C}$ is the generate loss of image generator, $\mathcal{L}_{D}$ is the adversarial loss.

During \textit{generation}, the input image $I$ also separated into cloth image $I_{c}$ and the rest part image $I_{b}$ just like in the training process. However, we only use the $I_{b}$ in the generation process. As for the cloth code, we randomly sampling a cloth part image $I^{'}_{c}$ of another person's image, and use the $\mathit{E_{A}}$ to extract the cloth code $c^{'}_{c}$ of $I^{'}_{c}$. Different from the training process, the $I^{'}_{c}$ will not be flipped and cropped during the generation process. With the randomly sampled cloth code $c^{'}_{c}$ and the rest part image $I_{b}$, the image generator $\mathit{G_{A}}$ would generate person image with different apparel but keep the person's identity in $I$ unchanged. The generated image samples has shown in \textbf{Fig.\ref{fig:4}}.

\subsection{Apparel-invariant Feature Learning framework}
\label{sec:pre-train}
\begin{figure}[!tp]
\includegraphics[width=0.50\textwidth]{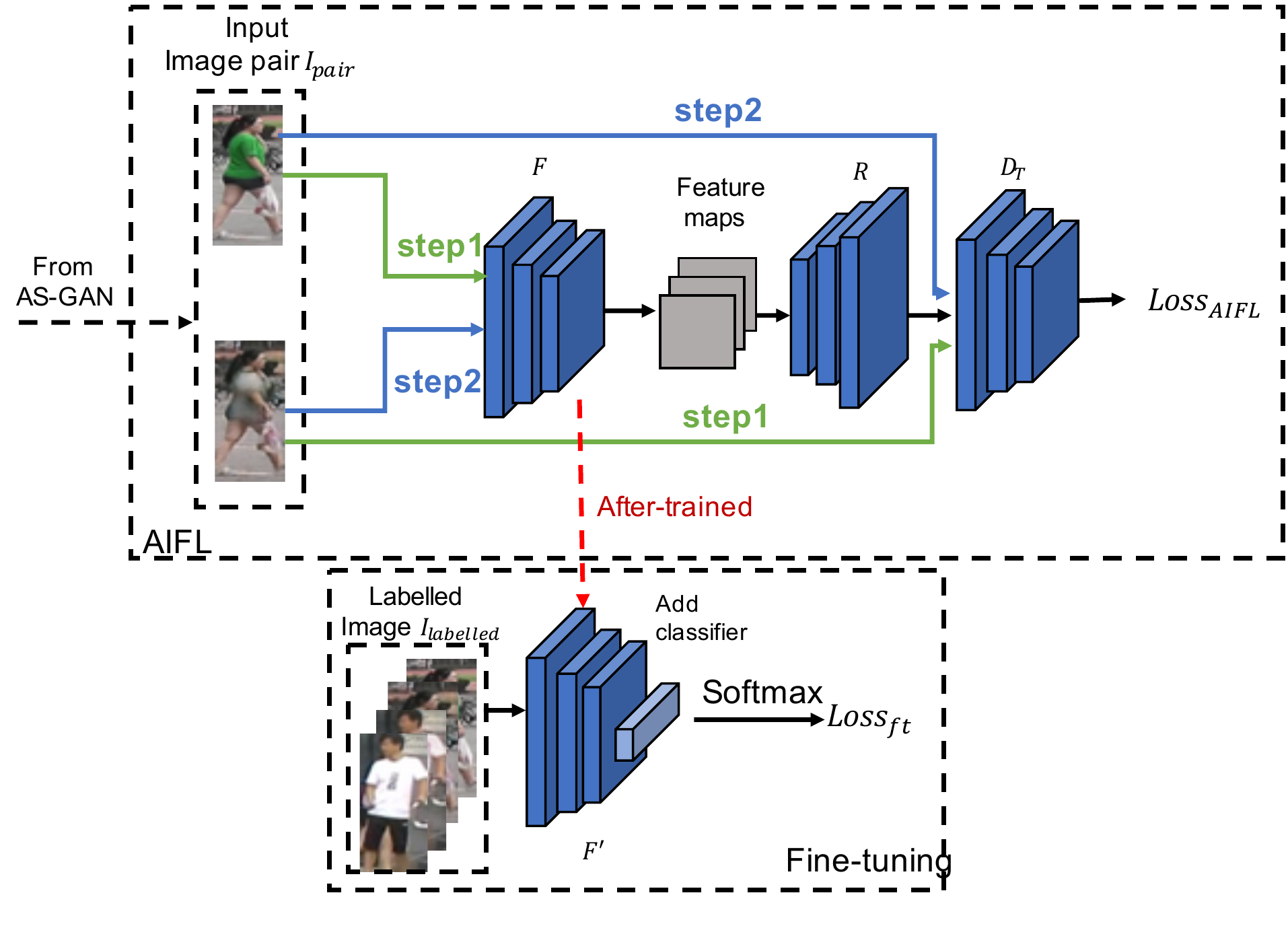}
\caption{\textbf{Framework} of our AIFL method. Our AIFL framework has an auto-encoder based structure. During training, each iteration is consists of two steps. In the first step, we set the original image as input and generated image as the target to run the whole network for once. In the second step, we set the generated image as input and the original image as the target. After the training process, we take out the feature extractor $F$ and fine-tuning it using softmax loss on the target dataset like MSMT17 and PAVIS.}
\label{fig:aiflframework}
\end{figure}

As shown in \textbf{Fig.\ref{fig:aiflframework}}, two key components construct the AIFL framework: an auto-encoder based transfer net $\mathit{T}$ and the followed discriminator $\mathit{D_{T}}$. This framework is trained in an unsupervised manner since we do not have any person or cloth label. 

\textbf{Apparel changed image generation}. Given a person image dataset, we first obtain the cloth code of each image in this dataset using our proposed apparel encoder $\mathit{E_{A}}$. Theoretically, we can obtain $n$ cloth code from a dataset contains $n$ images. Hence, we can generate $n-1$ apparel changed images for each person using the image generator $\mathit{G}$.

\textbf{Transfer net $\mathit{T}$}. 
To the best of our knowledge, no existing deep-learning based model has been proposed to solve the apparel changed person ReID problem. The main idea of our Apparel-invariant Feature Learning (AIFL) framework is to train a feature extractor to learn the apparel-irrelevant information. To achieve this goal, given an off-the-shelf CNN model $\mathit{F}$ like ResNet-50 or DenseNet-161, we first implant $\mathit{F}$ as the encoder in the transfer net $\mathit{T}$ and followed by a decoder $\mathit{R}$. To improve the performance, we also applied a discriminator $\mathit{D_{T}}$ to distinguish whether an input image is real. After the training process, the $\mathit{F}$ will be taken out and be fine-tuned into the finial ReID model using softmax loss. The structure of $\mathit{T}$ has shown in \textbf{Fig.\ref{fig:aiflframework}}. 

During \textit{training}, we first pair up the synthetic image $\widetilde{I}$ with the original image $I$. As shown in \textbf{Fig.\ref{fig:aiflframework}}, the transfer net will be trained in the step in each iteration. In step 1, we set the $I$ as input and $\widetilde{I}$ as the reconstruct target to train the whole transfer net $\mathit{T}$. We took out the feature maps of the last Conv2d layer of the feature extractor $\mathit{F}$ and fed into the decoder $\mathit{R}$. In the decoder, the feature maps will be used to reconstruct an image with the same size as the original image. Next, we calculate the loss given in \textbf{Eq.\ref{equ:auto-total}} and conduct backpropagation. After that, we reverse the input and target input to train the whole transfer net $\mathit{T}$ in step 2.

Since the synthetic image generated image and original image contain the same person worn different clothes, the model has to ignore the difference (apparel) between these two images to yield a lower loss in the training process. Moreover, the illumination variations of images containing the same person could also be mitigated by imperfect image generation. 
As for objective, we use a L1 loss as the reconstruction loss, as shown in \textbf{Eq.\ref{equ:loss-auto}}:
\begin{equation}
\label{equ:loss-auto}
\begin{split}
\mathcal{L}_{recon} = \parallel I_{target} -  \mathit{T}(x) \parallel_{1}
\end{split}
\end{equation}
Where x is the input image ($I$ in step 1 and $\widetilde{I}$ in step 2), and $I_{target}$ is the target image ($\widetilde{I}$ in step 1 and $I$ in step 2). 

\textbf{Discriminator $\mathit{D_{T}}$}. To improve the performance of our model, we construct a 7 layers discriminator $\mathit{D_{T}}$ including 6 Conv2d blocks and a linear classification layer. Each convoluational blocks contains a Conv2d layer and a BatchNorm layer and a LeakyReLU layer with negtive slope 0.2.  The discriminator is aims to differentiate the input images is real or fake by generate a binary classification mask.  The adversarial loss has shown below:

\begin{equation}
\label{equ:discriminator}
\begin{split}
\mathcal{L}_{\mathit{D_{T}}}= \min_{\mathit{D_{T}}}\max_{\mathit{T}(x)} \mathbb{E}_{x}\log[\mathit{D_{T}}(x)]+\mathbb{E}_{\mathit{T}(x)}\log[1-\mathit{D_{T}}(\mathit{T}(x)))]
\end{split}
\end{equation}
Where x is input image, $\mathit{T}(x)$ is reconstruction output of $\mathit{T}$ with input x.

This objective encourages the transfer net $\mathit{T}$ to reconstructed a more realistic image, and eventually makes the features extracted by $\mathit{F}$ of $\widetilde{I}$ and $\widetilde{I}$ closer. We also conducted an ablation experiment to demonstrate the effects of the discriminator, which will be demonstrated later. 

Overall, the joint loss of the AIFL framework can be expressed as \textbf{Eq.\ref{equ:auto-total}}:
\begin{equation}
\label{equ:auto-total}
\begin{split}
\mathcal{L}_{AIFL}= \mathcal{L}_{recon} + \lambda_{\mathit{D_{T}}}\mathcal{L}_{\mathit{D_{T}}}
\end{split}
\end{equation}
Where $\lambda_{\mathit{D_{T}}}$ is the hyper-parameter to control the impact of the discriminator, we select 0.001 by practice in all the following experiments.

\subsection{Fine-tuning algorithm}
After the training process in the Transfer net, we took out the trained CNN model $\mathit{\widetilde{F}}$ from $\mathit{T}$ and added an average pooling layer and three linear layers as a classifier at the end of the $\mathit{\widetilde{F}}$. After that, we conduct the fine-tuning process on $\mathit{\widetilde{F}}$ as in standard classification task using softmax loss. In the fine-tuning process, a small number of labeled images will be used to fine-tune this model. The label information here only contains person ID, and the cloth ID is not requested.

\begin{table*}
\caption{Summary of datasets}
\label{tab:datasets}
\begin{center}
\resizebox{\textwidth}{!}{
\begin{tabular}{c|ccccccc}
Dataset & Cams & Max. cloth change$^{1}$ & Training (Images/IDs) & Gallery (Images/IDs) & Query (Images/IDs) & Validation (Images/IDs) & Total (Images/IDs) \\ \hline
PAVIS$^{2}$   & 2    & 1                          & 36/-                & 35/-               & 35/-             & 8/-                    & 788/79             \\
MSMT17  & 15   & -                          & 30,248/1,041          & 82,161/3,060         & 11,659/3060        & 2,373/1,041             & 126,441/4,101      \\
C-MARS$^{3}$  & 6    & 6                          & 600,000               & -                    &   -                 & -                       & 600000            
\end{tabular}}
\begin{tablenotes}
\small
\item 1 The maximum cloth change times. The cloth id in the MSMT17 is not available.
\item 2 We split the PAVIS dataset 10 times. The image number of each set is not fixed. The gallery set IDs is the same as in the query set.
\item 3 The C-MARS contains 100,000 images sampled from the MARS dataset, and 5 sets of synthetic images generated by AS-GAN using the sampled images. These images only use to initialise the CNN model.
\end{tablenotes}    
\end{center}
\end{table*}

\section{Experiement and Results}
\label{sec:experiement}
\subsection{Datasets}
We conduct experiments on two ReID datasets to evaluate the performance of the proposed method, including PAVIS\cite{Barbosa:reid12} and MSMT17\cite{DBLP:journals/corr/abs-1711-08565}. The irrelevant person images used in AIFL and AS-GAN are sampled from a large-scale unlabeled MARS \cite{zheng2016mars} dataset. The summary of these datasets could be found in \textbf{Table \ref{tab:datasets}.}

\textbf{PAVIS}. We used two groups of images in the PAVIS dataset for evaluation here, denoted by 'Walking1' and 'Walking2'. Images of 'Walking1' and 'Walking2' were obtained by recording the same 79 people with a frontal view, walking slowly in an indoor scenario using an RGB-D camera. Among these 79 persons, from one group to another, 60 persons have changed their apparel. Each person has 4 or 5 and no more than 5 images in each group. By following the train-test policy, we randomly sampled 36 persons for training, 8 persons for validation, and 35 persons for testing. We repeated the dataset split 10 times and trained 3 models in each dataset. We then take the arithmetic mean of these model's evaluation results as the final result.

\textbf{MSMT17}. The cloth id for each person is not available in this dataset as well. However, the raw video in this dataset is recorded in 4 days with different weather conditions in a month using 12 outdoor cameras and 3 indoor cameras. Therefore, the illumination variations and pedestrian apparel changes in this dataset are richer. Meanwhile, there are abundant different persons wearing similar cloth cases in this dataset. The MSMT17 dataset contains 126,441 bounding boxes of 4,101 identities. The dataset partition has shown in \textbf{Table \ref{tab:datasets}}, we followed the same dataset split in \cite{DBLP:journals/corr/abs-1711-08565}, and used the evaluation code provided by them as well.

\textbf{C-MARS.} We randomly sampled 100,000 images without any label information from the large-scale dataset MARS\cite{zheng2016mars} as the training data in the AIFL framework. The images in MARS is collected by five HD cameras and one SD camera. There are more than 1.1 million bounding-boxes in this dataset. Moreover, it is consists of 1,261 different pedestrians who are captured by at least two cameras. We generated five sets of cloth changed images for the sampled images using AS-GAN to form the C-MARS dataset. As shown in \textbf{Table \ref{tab:datasets}}, there are 60,000 image in the C-MARS datasets. Each synthetic image of the five sets was paired up with the original images to form 50,000 image pairs. One set corresponds to 100,000 pairs of images. This dataset is only used to initialize the feature extractor model via the AIFL framework. Hence, we do not keep label information in this dataset.

\subsection{Evaluation Settings} 
\textbf{Feature extractor $\mathit{F}$}. Most of the existing CNN models could be implanted in the proposed AIFL framework as the feature extractor. In this work, we selected two off-the-shelf model, ResNet-50\cite{he2016deep} and DenseNet-161\cite{DBLP:journals/corr/HuangLW16a} to illustrate the effectiveness of our method. Due to the outstanding performance, these models are widely used in many computer vision tasks, including person ReID. All the ResNet-50 and DenseNet-161 models used in the experiments were pre-trained on the ImageNet\cite{imagenet_cvpr09}. Moreover, all of the codings are completed using Pytorch with version 0.4.0 and torchvision with version 0.2.1.

\textbf{Parameter setting.} All images used in this work are resized to 256 x 128 (height x width) in advance. All experiments used the same parameter settings. The batch size in the initializing process and the fine-tuning process are set to 128 and 64.  We use stochastic gradient descent (SGD) with 0.9 momentum and 5e-4 weight decay as an optimizer in all experiments. 

In the initializing process in the AIFL framework, we used an initial learning rate of 0.1, and decline 10 times per epoch. We set the 0.001 hyper-parameter $\lambda_{D_{T}}$ by practice.

In the fine-tuning process with softmax loss, the newly added layers use 0.1 as the initial learning rate, and the rest layers use 0.01 as the initial learning rate.

We conducted randomly cropping, resizing and horizontal flipping on training images for data augmentation during the fine-tuning process in all experiments. Unless otherwise specified, the model after the initializing process will be fine-tuned for 20 epochs. 
During the evaluation, we take out the output feature of the average pooling layer as the input image's embedding.
Every experiment is repeated for 3 times, and we only reported the mean values. 

\subsection{Evaluation Protocals}
We applied two evaluation methods to quantitatively measure the ReID performance:
\begin{enumerate}
\item mean Average Precision(mAP);
\item Cumulative Matching Characteristics (CMC).
\end{enumerate}

\begin{table}[!tp]
\centering
\caption{Results in PAVIS}
\label{tab:pavis}
\begin{minipage}{\columnwidth}
\begin{center}
\resizebox{\textwidth}{!}{
\begin{tabular}{c|ccc}
Network Structure       & mAP            & CMC@1          & CMC@5          \\ \hline
AlignedReID\cite{zhang2017alignedreid}             & 59.66          & 53.64          & 56.10          \\ \hline
ResNet-50 (baseline)    & 59.20          & 47.10          & 57.70          \\
DenseNet-161 (baseline) & 59.90          & 50.10          & 59.70          \\ \hline
ResNet-50 (Our)         & 59.90          & 53.20          & 57.40          \\
DenseNet-161 (Our)      & \textbf{64.50} & \textbf{59.40} & \textbf{61.00}
\end{tabular}}
\end{center}
\end{minipage}
\end{table}

\subsection{Evaluation on PAVIS}
We first use the image pairs in the C-MARS dataset to initialize the ResNet-50 and DenseNet-161 in the AIFL framework. After the initializing process, we took out the CNN model and fine-tuned it use the labeled images in the PAVIS dataset.

As for baselines, we compared with the DenseNet-161 and ResNet-50 model pre-train on ImageNet and also fine-tuned in the PAVIS dataset. We also compared with the other state-of-the-art person ReID method AlignedReID\cite{zhang2017alignedreid} on the PAVIS dataset. To the best of our knowledge, the author has not released its codes. Therefore, the implementation of AlignedReID we used is the reproduce version\footnote{https://github.com/huanghoujing/AlignedReID-Re-Production-Pytorch} on the Github. 

As we mentioned above, the PAVIS dataset is consists of two groups of images. People had changed their cloth when the other camera captured him. We have not compared with the previous work \cite{Barbosa:reid12} on the PAVIS because it is an RGB-D based method. Meanwhile, they used a specific experiment setting by using 'Walking 1' as the query set and 'Walking 2' as the gallery set. To evaluate our method in a more persuasive method, we used all the images of the test part IDs as query and gallery set in the same time. During the evaluation, we conduct the cross-group person ReID on the query set images, which means images in the same group will be ignored. Therefore, the model needs to re-identify the images in 'Walking 1' and re-identify the images in 'Walking 2' by using the 'Walking 1' as gallery set. 

\textbf{Results} has shown in \textbf{Table \ref{tab:pavis}}. We can notice that initializing the feature extractor model using the AIFL framework could significantly improve the performance. Both the ResNet-50 and DenseNet-161 model initialized via the AIFL framework outperforms the baseline models at higher matching accuracies in each rank. The performance of DenseNet-161 is raised from 59.9\% to 64.5\% on mAP.

\subsection{Evaluation on MSMT17}
As we mentioned above, the MSMT17 dataset does not have cloth IDs as well. However, the relatively long time interval has brought variations like illumination and apparel changes in this dataset. To verify the effect of our proposal on the feature extractor model in the apparel-invariant feature learning scenario. We followed the same evaluation setting as in \cite{DBLP:journals/corr/abs-1711-08565} to compare with the previous works. The dataset split has shown in \textbf{Table \ref{tab:datasets}}. We conduct the same initialize process as in the PAVIS dataset before fine-tuning it on the MSMT17 dataset. 

As for baseline, we use the ResNet-50 and DenseNet-161 pre-train on ImageNet as baseline and fine-tuned it using the train set of MSMT17. Meanwhile, we also compare with three state-of-the-art methods including GoogLeNet\cite{DBLP:journals/corr/SzegedyLJSRAEVR14}, PDC\cite{su2017pose} and GLAD\cite{DBLP:journals/corr/abs-1709-04329}.

\begin{table}[!htp]
\centering
\caption{Results in MSMT17}
\label{tab:msmt17}
\begin{minipage}{\columnwidth}
\begin{center}
\resizebox{\textwidth}{!}{
\begin{tabular}{c|cccccc}
Network Structure       & Training data  & mAP            & CMC@1          & CMC@5          & CMC@10         & CMC@20         \\ \hline
GoogLeNet               & MSMT17         & 23.00          & 47.60          & 65.00          & 71.80          & 78.20          \\
PDC                     & MSMT17         & 29.70          & 58.00          & 73.60          & 79.40          & 84.50          \\
GLAD                    & MSMT17         & 34.00          & 61.40          & 76.80          & 81.60          & 85.90          \\ \hline
ResNet-50 (baseline)    & MSMT17         & 26.32          & 55.21          & 70.61          & 76.48          & 81.61          \\
DenseNet-161 (baseline) & MSMT17(extend)$^{1}$ & 21.42          & 49.44          & 64.83          & 71.24          & 77.25          \\
DenseNet-161 (baseline) & MSMT17         & 29.43          & 61.67          & 74.77          & 79.49          & 83.36          \\ \hline
ResNet-50 (our)         & MSMT17         & 29.66          & 58.28          & 73.33          & 79.15          & 83.79          \\
DenseNet-161 (our)      & MSMT17(extend) & 22.34          & 49.68          & 65.74          & 72.13          & 78.27          \\
DenseNet-161 (our)      & MSMT17         & \textbf{35.24} & \textbf{65.96} & \textbf{79.01} & \textbf{83.80} & \textbf{87.76}
\end{tabular}}
\end{center}
\end{minipage}
\begin{tablenotes}
\item 1. MSMT17(extend) is consists of the original training set of MSMT17 and the 3 sets of synthetic images generated by AS-GAN using the training set image of MSMT17. The generated images is labelled with the same person ID and camera ID as in the original image. 
\end{tablenotes}
\end{table}

We summarize the experimental Results in \textbf{Table \ref{tab:msmt17}}. 
As shown in the table, the DenseNet-161 model initialized by our proposal substantially outperforms all the baseline's performance. We also notice that the models initialized via the AIFL framework could achieve a better person ReID performance in the MSMT17 dataset. The above experiments clearly show the effectiveness of our proposal in learning an apparel-invariant feature.
As we mentioned above, the most direct way to use the generated images is to extend training data. Therefore, we generated 3 sets of synthetic images for each image in the training set of MSMT17. We use these synthetic images and the original images in the MSMT17 training set to form an extended training set denote as MSMT17(extend).

The results has shown in \textbf{Table \ref{tab:msmt17}}. We can notice that directly use these synthetic images is not helpful in the training process. We can observe a tremendous performance fallout in both the baseline and the model initialized by the AIFL. However, the model initialized using the AIFL still outperforms the baseline model.

\subsection{Further Evaluations}
\begin{table}[!tp]
\caption{Evaluate the impact of the refined layer}
\label{tab:norefined}
\begin{minipage}{\columnwidth}
\begin{center}
\resizebox{\textwidth}{!}{
\begin{tabular}{c|ccccc}
Network Structure      & mAP   & CMC@1 & CMC@5 & CMC@10 & CMC@20 \\ \hline 
DenseNet-161 (normal)$^{1}$  & 33.10 & 65.63 & 78.36 & 82.58  & 86.67  \\
DenseNet-161 (refined)$^{2}$ & 35.24 & 65.96 & 79.01 & 83.80  & 87.76 
\end{tabular}}
\end{center}
\end{minipage}
\begin{tablenotes}
\item 1 This model is initialized using the image pairs generated by the AS-GAN with normal Conv2d blocks.
\item 2 This model is initialized using the image pairs generated by the AS-GAN with the refined layer.
\end{tablenotes}
\end{table}
\textbf{\textit{Evaluate the impact of the refined layer}}. As we mentioned above, we used the refined layer to replace the second normal Conv2d block in the AS-GAN. This modify could impact the generated image, and further impact the initialising process of the AIFL framework. To evaluate the influence of this modify, we conduct several experiments in the MSMT17 dataset. In these experiments, we use the normal Conv2d block as the secend layer in the AS-GAN to trained a image generator as baseline. We then generated 5 sets of image pairs using this image generator. With these image pairs, we initialised a DenseNet-161 model by following the same setting in the MSMT17 experiments. The generate image samples has shown in \textbf{Fig.\ref{fig:4}}, the left part is the image samples generated using AS-GAN with refined layer and the right part is image samples generated using normal Conv2d block. We can observe that the left part images are more realistic than the right part. As for numeric metric, the results has shown in \textbf{Table. \ref{tab:norefined}}, we can notice that by using the refined layer in the AS-GAN, we could generate more realistic image and further improve the performance of the AIFL framework.

\begin{table}[!tp]
\caption{Data Volume Increase Experiment}
\label{tab:data-volume}
\begin{minipage}{\columnwidth}
\begin{center}
\resizebox{\textwidth}{!}{
\begin{tabular}{c|cccccc}
Network Structure   & Data Volume  & mAP   & CMC@1 & CMC@5 & CMC@10 & CMC@20 \\ \hline
DenseNet-161 (AIFL) & 1 sets & 30.80 & 61.64 & 75.82 & 80.97  & 85.25  \\
DenseNet-161 (AIFL) & 3 sets$^{1}$       & 32.04 & 64.07 & 76.88 & 81.59  & 85.73  \\
DenseNet-161 (AIFL) & 5 sets       & 35.24 & 65.96 & 79.01 & 83.80  & 87.76 
\end{tabular}}
\end{center}
\end{minipage}
\begin{tablenotes}
\item 1. The 3 sets in the data volume means we randomly selected 60\% of the 500,000 generated images and paired up with the original images to form 300,000 image pairs. One set corresponds to 100,000 pairs of images.
\end{tablenotes}
\end{table}

\textbf{\textit{Data volume increase experiment}}. In our assumption, the AIFL framework could achieve better performance by using more image pairs. To verify this assumption, we conducted the data volume increase experiment using the C-MARS dataset. We conduct two groups of experiments. In each group, we randomly select 1 and 3 sets of image pairs from the C-MARS dataset to initialize the feature extractor model using the AIFL framework. To evaluate the performance, we also used MSMT17 to fine-tune and test the feature extractor model's performance. The results have shown in \textbf{Table \ref{tab:data-volume}}. As in the table, using more image pairs will increase the effectiveness of the AIFL framework. As an unsupervised training framework, the obtaining of the training data is relatively easier. Meanwhile, the training data we used is randomly sampled from irrelevant datasets, making the framework more practical and easy to implement in different datasets. These results also verified our contribution in this work that our proposal could effectively improve the apparel changed ReID performance by generated realistic images.

\textbf{\textit{Ablation}}. To verify the necessity of the discriminator $\mathit{D_{T}}$ we used in the AIFL framework, we conduct the ablation experiment in the MSMT17 dataset.
We summarise the experimental Results \textbf{Table. \ref{tab:ablation}}. We can observe a considerable decline in both mAP and CMC score when initializing the model without the discriminator $\mathit{D_{T}}$. We argue that the discriminator could help the model to learn a more accurate feature representation in the AIFL framework.

\begin{table}[!tp]
\caption{Discriminator Ablation Experiment.}
\label{tab:ablation}
\begin{minipage}{\columnwidth}
\begin{center}
\resizebox{\textwidth}{!}{
\begin{tabular}{c|ccccc}
Network Structure    & mAP   & CMC@1 & CMC@5 & CMC@10 & CMC@20 \\ \hline
DenseNet-161 (AIFL $no \mathit{D_{T}}$)$^{1}$ & 31.61 & 62.50 & 76.31 & 81.13  & 85.44  \\
DenseNet-161 (AIFL)  & 35.24 & 65.96 & 79.01 & 83.80  & 87.76 
\end{tabular}}
\end{center}
\end{minipage}
\begin{tablenotes}
\item 1. $no \mathit{D_{T}}$ stands for the model initialized using the AIFL framework but without discriminator $\mathit{D_{T}}$.
\end{tablenotes}
\end{table}
\section{Conclusion}
In this paper, we study the apparel changing and different persons wearing similar cloth cases in person ReID tasks. To learn an apparel-invariant feature embeeding, we proposed a semi-supervised apparel-invariant feature Learning (AIFL) framework. We also constructed an apparel-simulation GAN (AS-GAN) to generate realistic cloth-changed images or similar person's images for AIFL framework. We show that an apparel-invariant feature could be learned via our approach. Nevertheless, only limited labelled images are requested in our approach, which could effectively reduce the cost of human labelling. Extensive experiments are conducted in two person ReID datasets shown the progressiveness of our proposal.

\textbf{Further Works}. As in the newly defined apparel-changed person ReID scenario, many possible ways could be explored in the future. In this work, the fine-tuning process is still a supervised method. We are looking forward to exploring a fully unsupervised method in this new scenario. 


\bibliographystyle{IEEEtran}
\bibliography{IEEEabrv,reference.bib}
\vfill

\end{document}